\documentclass[letterpaper, 10 pt, conference]{ieeeconf} 
\IEEEoverridecommandlockouts 
\title{\LARGE \bf
Perch a quadrotor on planes by the ceiling effect
}

\author{Yuying Zou, Haotian Li, Yunfan Ren, Wei Xu, Yihang Li, Yixi Cai, Shenji Zhou and Fu Zhang
\thanks
{All authors are with the Department of Mechanical Engineering, The University of Hong Kong (HKU), Pokfulam, Hong Kong. \{\textit{zyycici, haotianl, renyf, xuweii, yhangli, yixicai, shenjizhou}\}\textit{@connect.hku.hk}, \{\textit{fuzhang}\} \textit{@hku.hk}. }
}

\usepackage{epsfig} % for postscript graphics files
\usepackage{times} % assumes new font selection scheme installed
\usepackage{amsmath} % assumes amsmath package installed
\usepackage{amssymb} % assumes amsmath package installed
\usepackage{amsmath,amssymb,amsopn,amstext,amsfonts}
\usepackage{bm}
\usepackage{float}
\usepackage{booktabs}
\usepackage{cite} %for realize [1]-[3]
\usepackage{gensymb}
\usepackage[dvipsnames]{xcolor}
\usepackage{blindtext}
\usepackage{hyperref}
\begin{document}

\maketitle
\pagestyle{empty}  % no page number for the second and the later pages
\thispagestyle{empty} % no page number for the first page
\thispagestyle{plain}
\pagestyle{plain}

%%%%%%%%%%%%%%%%%%%%%%%%%%%%%%%%%%%%%%%%%%%%%%%%%%%%%%%%%%%%%%%%%%%%%%%%%%%%%%%%
\begin{abstract}

Perching is a promising solution for a small unmanned aerial vehicle (UAV) to save energy and extend operation time. This paper proposes a quadrotor that can perch on planar structures using the ceiling effect. Compared with the existing work, this perching method does not require any claws, hooks, or adhesive pads, leading to a simpler system design. This method does not limit the perching by surface angle or material either. The design of the quadrotor that only uses its propeller guards for surface contact is presented in this paper. We also discussed the automatic perching strategy including trajectory generation and power management. Experiments are conducted to verify that the approach is practical and the UAV can perch on planes with different angles. Energy consumption in the perching state is assessed, showing that more than 30\% of power can be saved. Meanwhile, the quadrotor exhibits improved stability while perching compared to when it is hovering.

\end{abstract}

%%%%%%%%%%%%%%%%%%%%%%%%%%%%%%%%%%%%%%%%%%%%%%%%%%%%%%%%%%%%%%%%%%%%%%%%%%%%%%%%
\section{Introduction}

Unmanned aerial vehicles (UAVs) have numerous applications such as aerial photography, surveying, and monitoring. However, UAVs suffer from certain constraints, and one of the most significant challenges is the limited flight time, which leads to reduced mission efficiency. A feasible solution to overcome this limitation is to develop the ability for UAVs to perch on environmental structures. Research has shown perching a drone can significantly reduce energy consumption, enhance drone stability, and has the potential for new applications \cite{hang2019perching, hsiao2019ceiling,bai2021design}.

Perching is a common behavior observed in birds and insects, where they land and rest on natural objects using their feet or claws, conserving energy and stabilizing their position. In the case of UAVs, many perching mechanisms such as grippers and hooks are bio-inspired \cite{hang2019perching, bai2021design, doyle2011avian, doyle2012avian,burroughs2016sarrus,roderick2021bird,zheng2023metamorphic}. These mechanisms can be either actively controlled by servo motors or passively actuated by the weight of the UAVs. Most of them are applicable for perching on branches, ropes, and fences. In the meantime, ongoing research is exploring strategies for perching on planar structures such as walls, buildings, and bridges, which are commonly encountered in urban environments. In the work of \cite{anderson2009sticky, daler2013perching, hawkes2013dynamic, kalantari2015autonomous, thomas2016aggressive}, adhesive pads are used to assist the UAVs to perch on walls temporarily. Mellinger \textit{et al.} \cite{mellinger2012trajectory} proposed the use of Velcro to attach UAVs to specific inclined surfaces, while Ji \textit{et al.} \cite{ji2022real} utilized magnets to apply pressure for perching on iron surfaces. 
Nonetheless, the above solutions do have some limitations. Firstly, incorporating additional mechanisms including but not limited to grippers and adhesive pads adds complexity to the design and extra weight to the UAVs. Although energy consumption can be reduced during perching, more power will be consumed during hovering and flight. Secondly, the mechanisms are mainly installed on the bottom side of the UAVs, which is typically where cameras are mounted for aerial photography. Those grippers or pads are not only constraining the views of the cameras but also potentially causing mechanical interference with the sensors. Last but not least, in all of the above methods, the UAVs must turn their bottom side towards branches or walls, which further obstructs the camera views leading to undesired filming results or even mission suspension. 

Recently, Hsiao and Chirarattananon \cite{hsiao2019ceiling} proposed a novel method of perching small rotorcraft by utilizing the ceiling effect \cite{powers2013influence, nishio2020stable,carter2021influence}. When UAVs approach to ceiling, the ceiling effect is an aerodynamic phenomenon creating a relatively low-pressure area between the ceiling and UAVs, which attracts rotors towards the surface, making them capable of perching. Moreover, this effect also reduces the drag on propellers, leading to higher rotating speeds and increased thrust. Experiments of perching a quadrotor under bridges were presented in \cite{sanchez2017multirotor, jimenez2019contact}, where the quadrotor was able to maintain altitude with lower throttle input. This method requires minimum mechanisms for UAVs to perch and barely affect the flight mission. 

\begin{figure}[t]
 \begin{center}
 {\includegraphics[width=1\columnwidth]{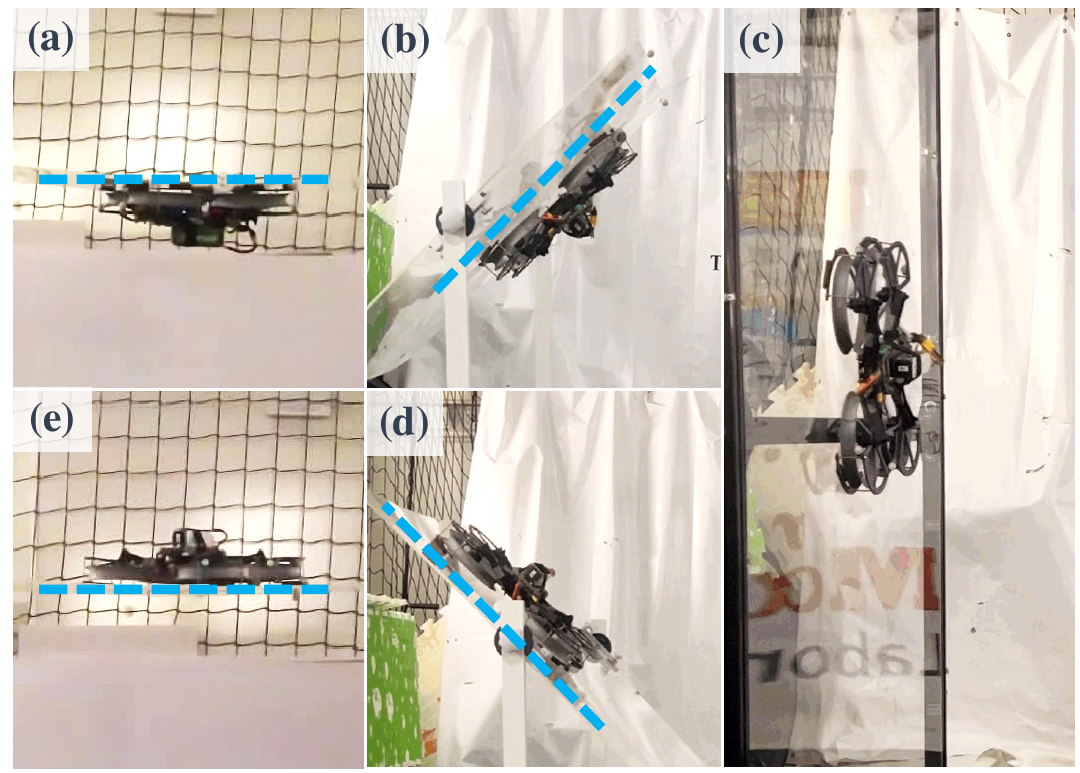}}
 \end{center}
 \caption{A quadrotor perching on planes with different incline angles including (a) 0\degree, (b) 45\degree, (c) 90\degree, (d) 135\degree, (e) 180\degree. The accompanying video is available at \url{https://youtu.be/6NZUERPpmFc}}
 \label{fig:angle}
\end{figure}

\begin{figure*}[t]
 \begin{center}
 {\includegraphics[width=1.8\columnwidth]{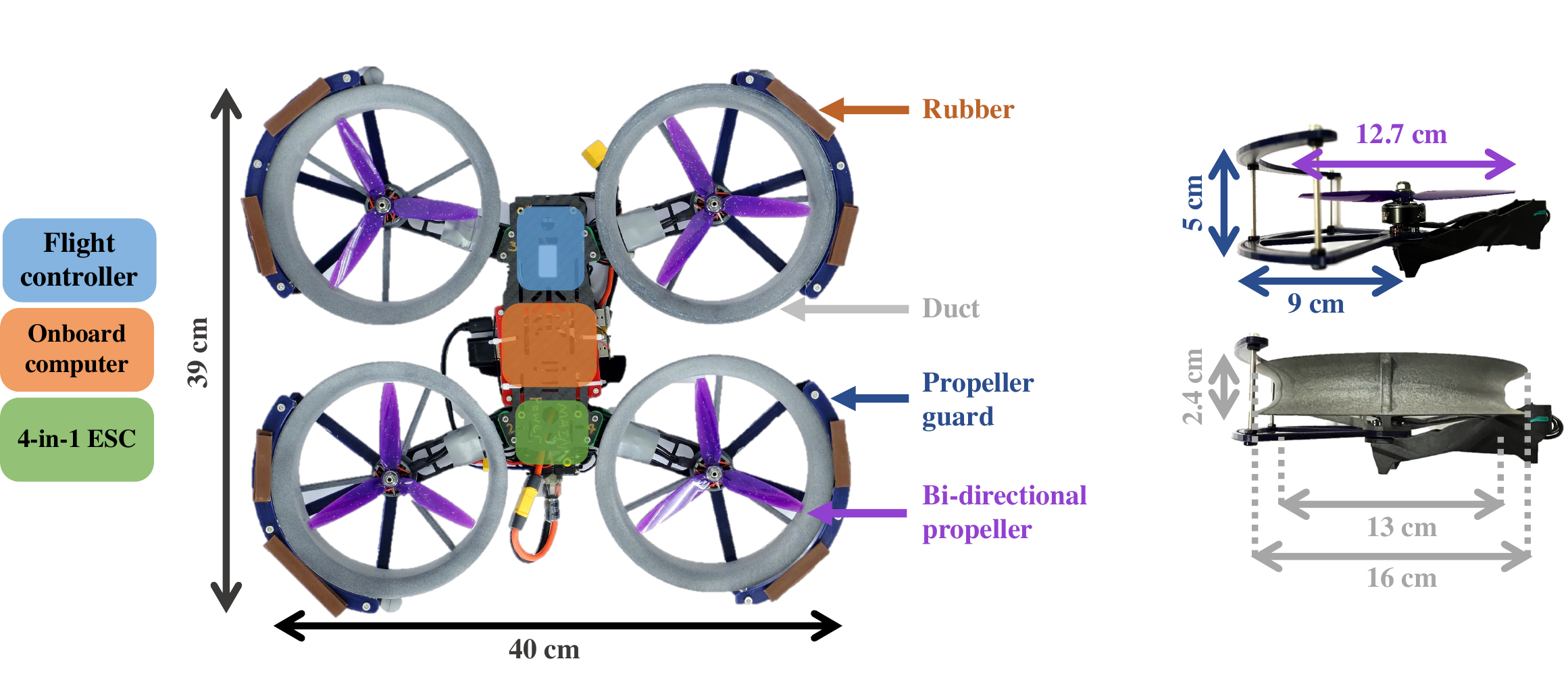}}
 \end{center}
 \caption{The structure and size illustration of the perching quadrotor.}
 \label{fig:component}
\end{figure*}

Similarly, when a quadrotor is closing to planar structures other than a ceiling, the quadrotor can experience the ceiling effect if it is aligned parallel to the surfaces. Therefore, we propose to perch a quadrotor on planes of varying incline angles including ceilings, walls, slopes, and grounds, as shown in Fig. \ref{fig:angle}. Only propeller guards are required as supporting structures when contacting with planar structures, which already exist on many drones for safety purposes. The quadrotor can save energy by perching on more available places. Besides, a distinctive landing strategy is introduced by this concept, which is flipping the quadrotor upside down and using the propeller guards as the landing gear. As traditional landing gears can be eliminated, the composition will be even simpler, and onboard cameras will have larger ranges of view in both perching and flight. 

One of the challenges is controlling the quadrotor to reach and maintain abnormal attitudes. Although modern electronic speed controllers (ESCs) enable bi-directional thrust generation in flight by changing the motor's rotation direction (known as 3D mode) \cite{maier2018bidirectional, jothiraj2019enabling, bass2020improving, jothiraj2020control, yu2020perching}, which allows the UAVs tilt to upright and upside-down postures as well as make firm contact on inclined surfaces, current research on perching trajectories mainly focuses on reaching the target position with the bottom side of the quadrotor and using single direction thrust \cite{zhang2013bio, ji2022real}. Plans for approaching the surfaces with the top side and using bi-directional thrust are rarely investigated in existing works. Thus, we present a coherent trajectory generation and control framework to adjust this challenge. 

In spite of several pieces of research indicating that the ceiling effect has the potential to save energy, the actual power that can be preserved on a complete quadrotor has not been assessed yet. The evaluation should be conducted while the quadrotor is stably perching with as less as possible thrust. Since we demonstrate a throttle control logic for perching, the power consumption in different conditions can be found in experiments.

We summarize the main contributions of this paper as follows:

\begin{enumerate}
 \item We design a quadrotor that can utilize the ceiling effect to perch on planar structures with its top side.
 \item We develop novel perching procedures for the quadrotor to reach different incline angles and verify the feasibility through experiments. 
 \item We evaluate the energy efficiency and stability of our perching method.
\end{enumerate}

\section{Methodology}
\begin{figure*}[t]
 \begin{center}
 {\includegraphics[width=1.8\columnwidth]{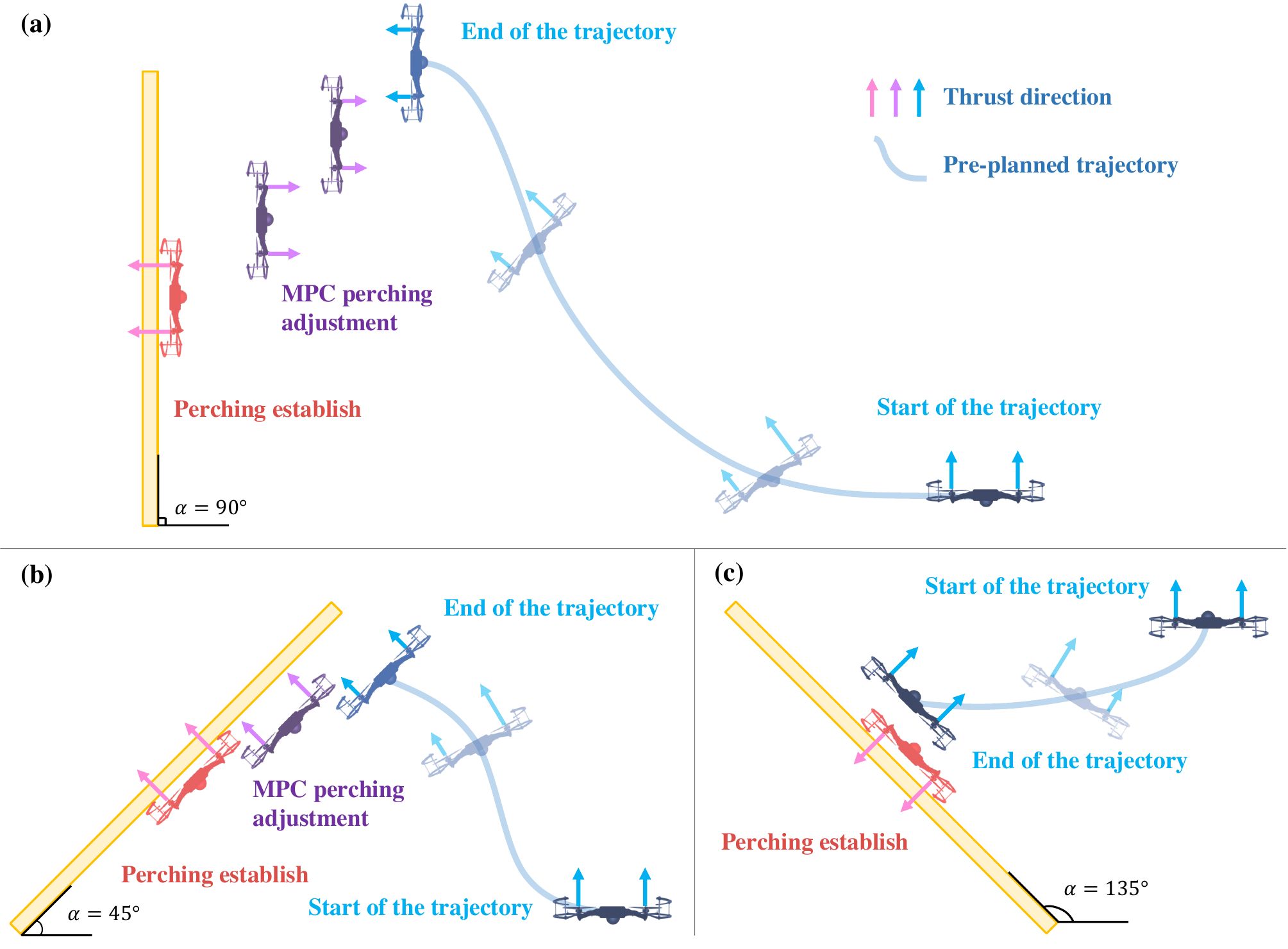}}
 \end{center}
 \caption{Examples of the perching process for planes with different incline angles. (a) Perch to a 90\degree{} wall. (b) Perch to a 45\degree{} inclined plane. (c) Perch to a 135\degree{} slope.}
 \label{fig:method}
\end{figure*}
\subsection{Design and control}

In this section, we first discuss the components and dimensions of the quadrotor as shown in Fig. \ref{fig:component}. Then, we elaborate on how the controller work for managing bi-directional thrust in general flight.
\subsubsection{Structure design}
This quadrotor is assembled from a purchasable regular fuselage, on which the propulsion system including four T-MOTOR F60 pro motors and GEMFAN 513D propellers are also common in the market. A T-MOTOR F60A 4-in-1 ESC with DShot1200 communication protocol is connected to all motors. Unlike variable-pitch propellers that produce reversible thrusts \cite{cutler2012actuator}, our method of using symmetrical-blade propellers and open-source ESC keeps the mechanism simple, ensuring effective bi-directional thrust while maximizing compatibility with existing UAVs.

Customized propeller guards are made of two 3D-print parts and long screws. The forces for supporting the quadrotor during surface contact are borne by the screws. A few rubber tapes are added on the top of the propeller guards to increase friction for perching on inclined planes as well as dampen the collision force. Light nylon ducts are used on this quadrotor to improve propulsion efficiency and enhance safety, by surrounding the propellers and enlarging the airflow velocity difference around them \cite{ai2021aerodynamic, li2021effect}.

Other components which include a flight controller Pixhawk 4 mini and an onboard computer Jasper Lake N5105 are attached at the center of the quadrotor's body. A 6S 1850 mAh battery weighted 270 \text{g} is mounted beneath the fuselage. The total weight of the quadrotor is 1131 \text{g} and the thrust-to-weight ratio is 2.5.

\subsubsection{System control}
This quadrotor is using a cascaded control framework base on PX4 that is also widely used in existing UAVs. The position and velocity controller produces the desired acceleration, which provides the desired thrust and attitude for the attitude controller and subsequent control loops. In order to reduce computational demands and complexity, we impose a constraint that the thrust generated by each propeller is always in the same direction, either all pointing upward or all pointing downward at the same time with respect to the quadrotor body. As a result, for one desired acceleration, there are two sets of desired thrust and attitude, including positive thrust with normal attitude and negative thrust with reverse attitude. The quadrotor only chooses the desired attitude and the relative thrust that is closer to its current attitude, preventing unnecessary flips. 

\subsection{Perching trajectory and strategy}

This section comprehensively introduces our automatic perching procedures for different cases, including \textbf{Case (a)} perching on the planes closing to vertical, which incline angle ranging from 60\degree{} to 120\degree{}, \textbf{Case (b)} perching on ceilings that have incline angle less than 60\degree{}, and \textbf{Case (c)} perching on grounds with incline angle greater than 120\degree{}. An example for each scenario is presented in Fig. \ref{fig:method}. We divide the whole perching process into steps, in every one of which the thrust directions of all propellers are unified and consistent. \textbf{Case (a)} requires three steps, starting with trajectory tracking only using positive thrust, followed by model predictive control (MPC) perching adjustment with reversed thrust, and then establishing and maintaining perching with positive thrust. \textbf{Case (b)} is the same except that positive thrust is used in MPC perching adjustment. \textbf{Case (c)} involves trajectory tracking with reversed thrust and perching establishing with positive thrust.

\begin{figure}[t]
 \begin{center}
 {\includegraphics[width=1\columnwidth]{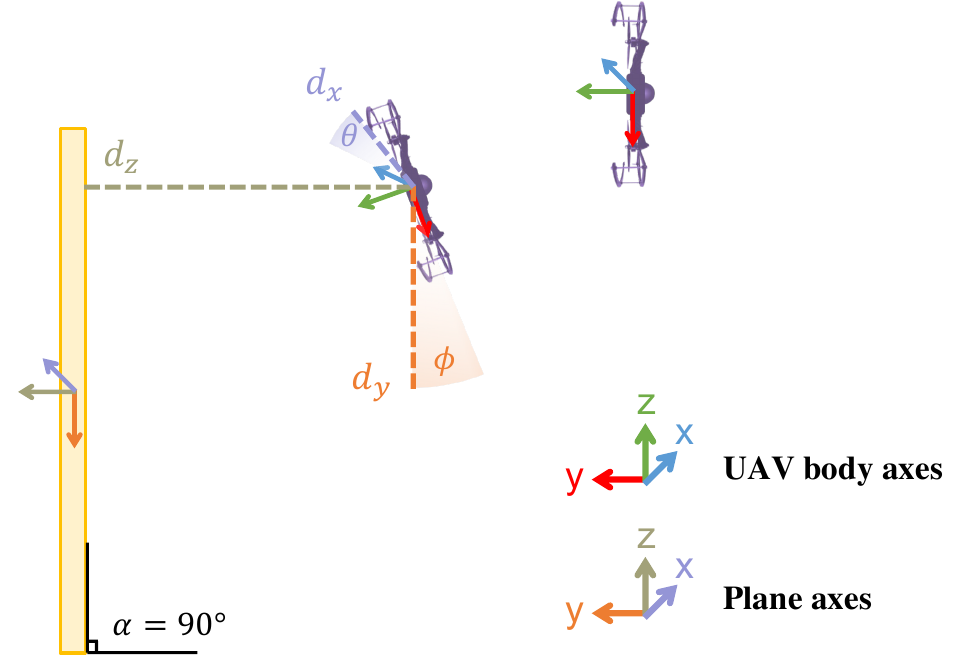}}
 \end{center}
 \caption{The definitions of key notations for perching procedures.}
 \label{fig:mpc}
\end{figure}

\subsubsection{Perching trajectory generation}
The first step of perching is to tilt the quadrotor from hovering to be parallel to the target surface. Since the thrust direction is fixed in this process for all cases, one of the state-of-the-art trajectory optimizers MINCO \cite{wang2022geometrically} can be employed. Our controller, as mentioned above, allows the quadrotor to choose reversed thrust and attitude based on the desired acceleration input in \textbf{Case (c)}. 

The initial state and the final state of the quadrotor are the input to the optimizer, while a smooth trajectory with constraints on mechanical properties is the output. Considering the plane axes are at the center of a surface and rotated accordingly (shown in Fig. \ref{fig:mpc}), the initial hover position is always in the $-z$ direction. The final position of the trajectory has offsets with regard to the center of the plane. As the quadrotor is tilted, its control effectiveness in the plane's $y$ direction is weakened and the motion is mainly dominated by the component of gravity in this direction. Space needs to be reserved for subsequent steps in view of this.

\subsubsection{MPC perching adjustment}
At the end of the trajectory tracking step, the quadrotor is parallel to the perching surface but is not ready to perch yet. On one hand, there are tracking errors, especially in \textbf{Case (a)} and \textbf{Case (b)} which have aggressive trajectories. The quadrotor position may not be ideal for perching. On the other hand, since the thrust points toward the plane in both \textbf{Case (a)} and \textbf{Case (b)}, there may be significant velocity in the plane's $+z$ direction, making a deceleration procedure necessary. Therefore, we set up the MPC perching adjustment to actively guide the quadrotor closing the surface for these two cases.

In this control section, the state vector comprises the quadrotor position with respect to the plane center, $d_z$, $d_y$, and $d_x$ as shown in Fig. \ref{fig:mpc}, and the relative velocity $v_z$, $v_y$, and $v_x$:

\begin{equation}\label{e:state}
{\textbf{x}}=
\left[ \begin{matrix}
d_z & v_z & d_y & v_y & d_x & v_x
\end{matrix}\right]^T.
\end{equation}

The control input consists of the thrust acceleration $a_T$, the relative roll angle $\phi$, and pitch angle $\theta$ (also indicated in Fig. \ref{fig:mpc}):

\begin{equation}\label{e:input}
{\textbf{u}}=
\left[ \begin{matrix}
 \cos\theta\cos\phi\ a_T & \sin\phi\cos\theta a_T & -\sin\theta\ a_T 
 \end{matrix}\right]^T.
\end{equation}

With the gravitational acceleration $g$ and the incline angle of the plane $\alpha$, this model is written as 

\begin{equation}\label{e:model}
\begin{cases}
 \dot v_z = \cos\theta\cos\phi\ a_T-g\cos(\alpha)\\
 \dot v_y = \sin\phi\cos\theta\ a_T+g\sin(\alpha)\\
 \dot v_x = -\sin\theta\ a_T\\
\end{cases}.
\end{equation}

We consider that there are total $n$ computational cycles in this adjustment period and the cycle interval is $\Delta t$ seconds. The total time of the adjustment period is $n\Delta t$ seconds. After discretization, we have
\begin{equation}\label{e:predict}
\textbf{x}(k+1)=
\boldsymbol{A}
\textbf{x}(k) + \boldsymbol{B} 
\textbf{u}(k) + \boldsymbol{d}, \quad k \in \mathbb{N},
\end{equation}
where $\boldsymbol{A} \in \mathbb{R}^{6 \times 6}$, $\boldsymbol{B} \in \mathbb{R}^{6 \times 3}$, $\boldsymbol{d} \in \mathbb{R}^{6 \times 1}$ can be computed from (\ref{e:model}).

The last state that is predicted from the $k$ cycle can be represented as 
\begin{equation}\label{e:predictend}
\textbf{x}(n|k)=
\boldsymbol{A}^{n-k}
\textbf{x}(k) \\+
\displaystyle\sum_{j=0}^{n-k-1}
\boldsymbol{A}^{n-k-1-j}
(\boldsymbol{B} \textbf{u}(k+j|k)+\boldsymbol{d})
.
\end{equation}

We define that 
\begin{equation}\label{e:Uset}
\textbf{U}_k=
\left[ \begin{matrix}
 \textbf{u}(k|k)\\
 \textbf{u}(k+1|k)\\
 \dots \\
 \textbf{u}(n-1|k)
 \end{matrix}\right],
\end{equation}
and our problem becomes
\begin{align*}
 \min \boldsymbol{J} &= {\textbf{U}_k}^T \textbf{I} \textbf{U}_k ,\tag{7a}\\
 \\
 \text{s.t.}\ \ 
 & 0 \leq d_z(n|k) \leq d_{z_{max}} ,\tag{7b}\\
 & 0 \leq v_z(n|k) \leq v_{z_{max}} ,\tag{7c}\\
 & d_{y_{min}} \leq d_y(n|k) \leq d_{y_{max}} ,\tag{7d}\\
 & d_{x_{min}} \leq d_x(n|k) \leq d_{x_{max}} ,\tag{7e}\\
 & \phi (n-1|k) =0 ,\tag{7f}\\
 & \theta (n-1|k) =0 ,\tag{7g}\\
 & a_{T_{min}} \leq a_T(j) \leq a_{T_{max}},
 j \in \{ k, \dots, n-1\}
 ,\tag{7h}\\
 & \phi_{min} \leq \phi(j) \leq \phi_{max},
 j \in \{ k, \dots, n-1\}
 ,\tag{7i}\\
 & \theta_{min} \leq \theta(j) \leq \theta_{max},
 j \in \{ k, \dots, n-1\} 
 ,\tag{7j}\\
\end{align*}
where we are finding the minimum input for fulfilling all constraints, including hard constraints of $d_z$, $v_z$, $d_y$, and $d_x$ in the last state ((7b)-(7e)), hard constraints of $\phi$ and $\theta$ in the last input ((7f)-(7g)), and constraints of $a_T$, $\phi$ and $\theta$ in all inputs ((7h)-(7j)). $a_{T_{min}}$ and $a_{T_{max}}$ are both negative in \textbf{Case (a)}.

In every cycle, we only take the first group of the desired attitude and thrust as the command. By repeating this control process for $n$ cycles, the influence of the tracking error from the previous step should be minimized, allowing the quadrotor to approach the center of the plane with low perpendicular velocity. While there are no constraints on the velocity along the planar surface, it will eventually be eliminated by friction.

\subsubsection{Throttle control in perching}
In order to establish perching and maintain it with reduced power consumption, implement control over the throttle of the quadrotor and give zero angular rate commands. Once reaches the plane, all propellers generate high positive thrust to push the quadrotor against the surface for a short period of time. Following this, the throttle is gradually reduced as long as the quadrotor remains stable in contact with the plane. The throttle value that no longer holds the quadrotor position is marked as $T_{\min}$. 

If there exists a $T_{\min}$, the high throttle is once again provided to rebuild the contact until the quadrotor stabilizes. The throttle is then gradually reduced again, but this time it stops reducing at $T_\text{perch}=T_{\min}+0.05$. However, in \textbf{Case (c)}, $T_{\min}$ may not exist, and all rotors can be turned off during perching to further conserve power in such a case.

\section{Experiments}

\subsection{Perch on different planes}
\begin{figure}[t]
 \begin{center}
 {\includegraphics[width=1\columnwidth]{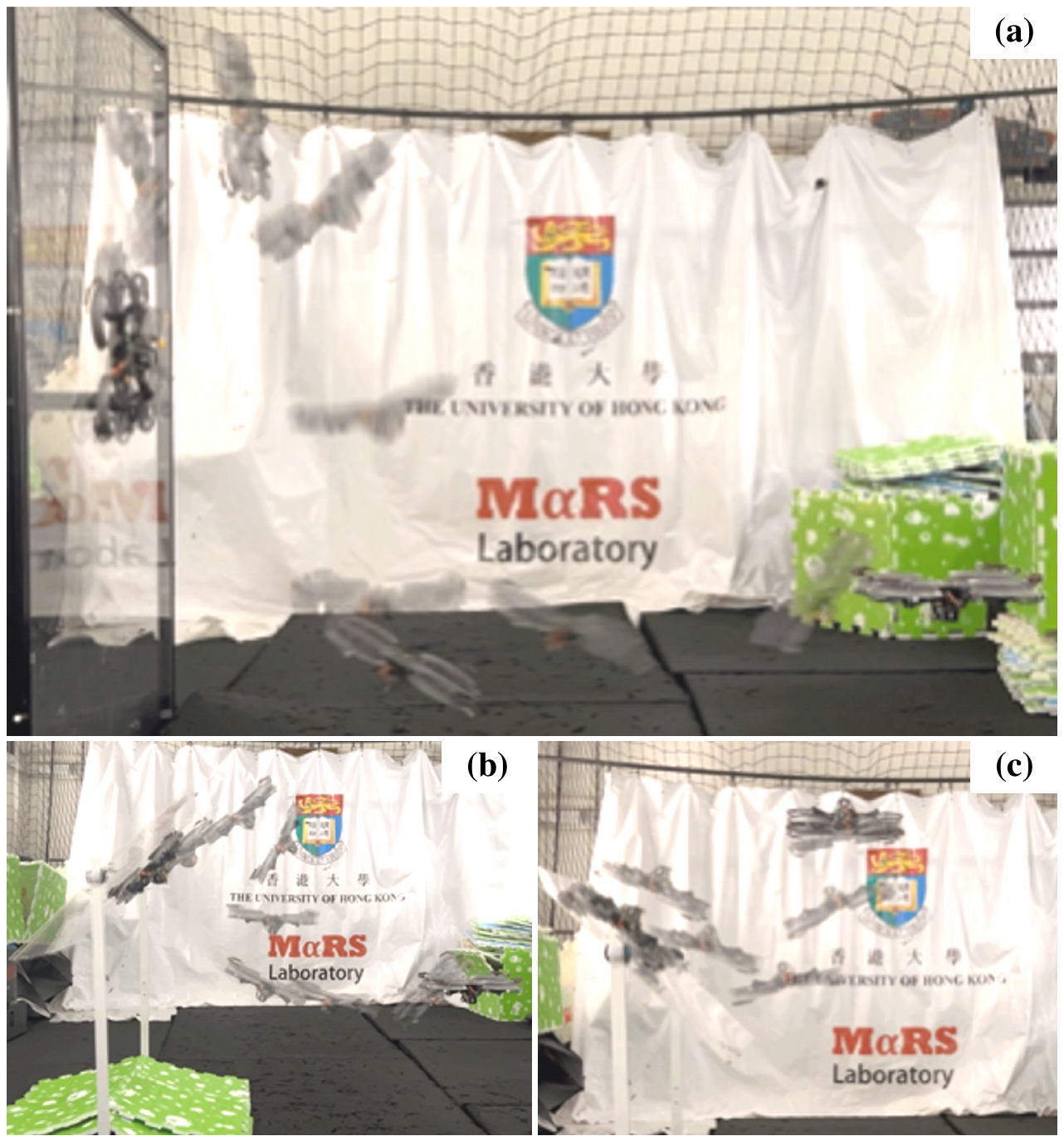}}
 \end{center}
 \caption{The processes of the quadrotor perching from hovering to three planes with (a) 90\degree{}, (b) 45\degree{}, (c) 135\degree{} incline angle respectively.}
 \label{fig:traj3}
\end{figure}
\begin{figure}[t]
 \begin{center}
 {\includegraphics[width=1\columnwidth]{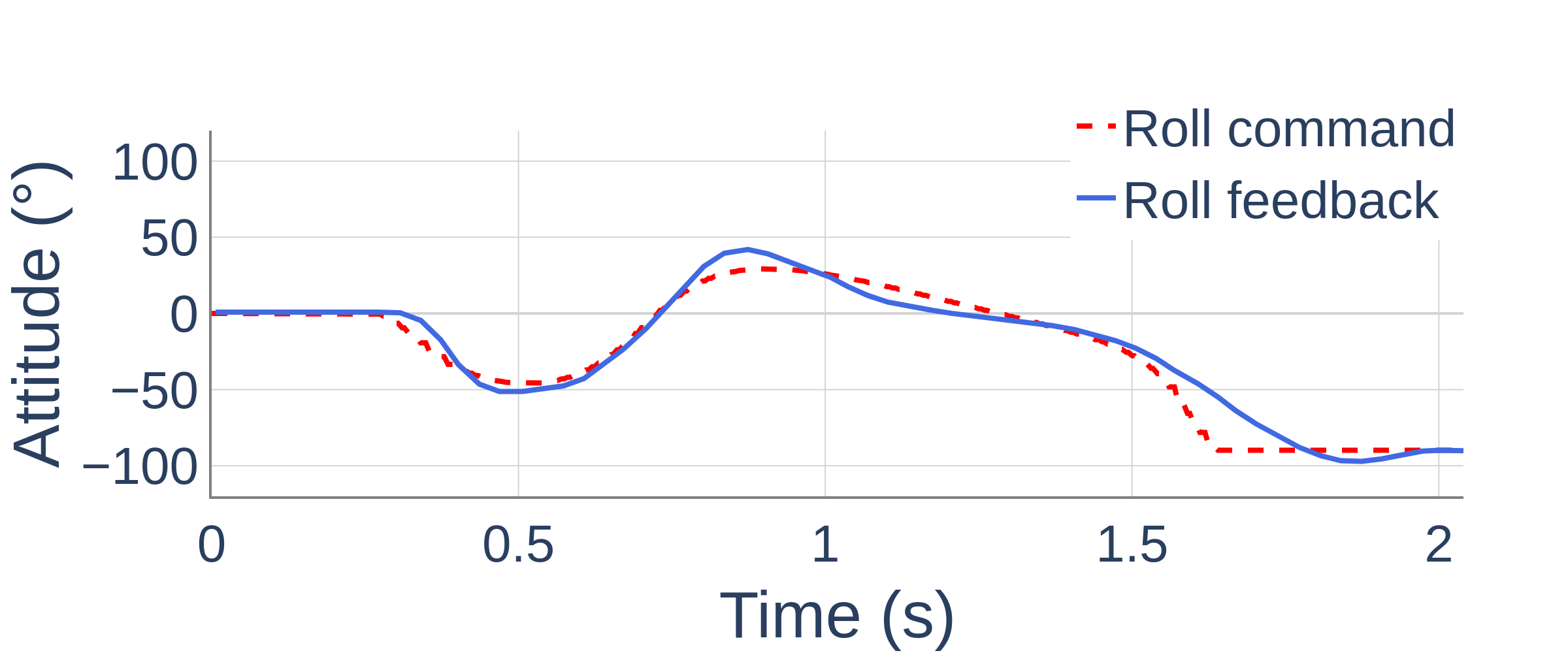}}
 \end{center}
 \caption{The roll command and its feedback in the experiment of perching to 90\degree{} wall.}
 \label{fig:roll}
\end{figure}

To verify the feasibility of our method, we carry out several tests of perching indoors. Transparent acrylic boards with a width of 120 cm are placed at different incline angles to represent various cases, as shown in Fig. \ref{fig:angle} and \ref{fig:traj3}. Both perching target planes and the quadrotor have pose feedback from the motion capture system. Fig. \ref{fig:traj3} also displays snapshots of the quadrotor in the perching processes and Fig. \ref{fig:roll}. depicts the roll command and feedback in the experiment of perching to a 90\degree{} wall. The quadrotor usually takes less than 2 seconds to reach the targeted angles. The experiments confirm that the quadrotor can tilt to be parallel to the planes and make contact successfully.

In addition, our results have shown to be highly repeatable. Despite the relatively smooth surfaces of the acrylic boards, the quadrotor successfully makes contact. Our strategy is proven to be fast, safe, and robust.

\subsection{Throttle control}
\begin{figure}[t]
 \begin{center}
 {\includegraphics[width=1\columnwidth]{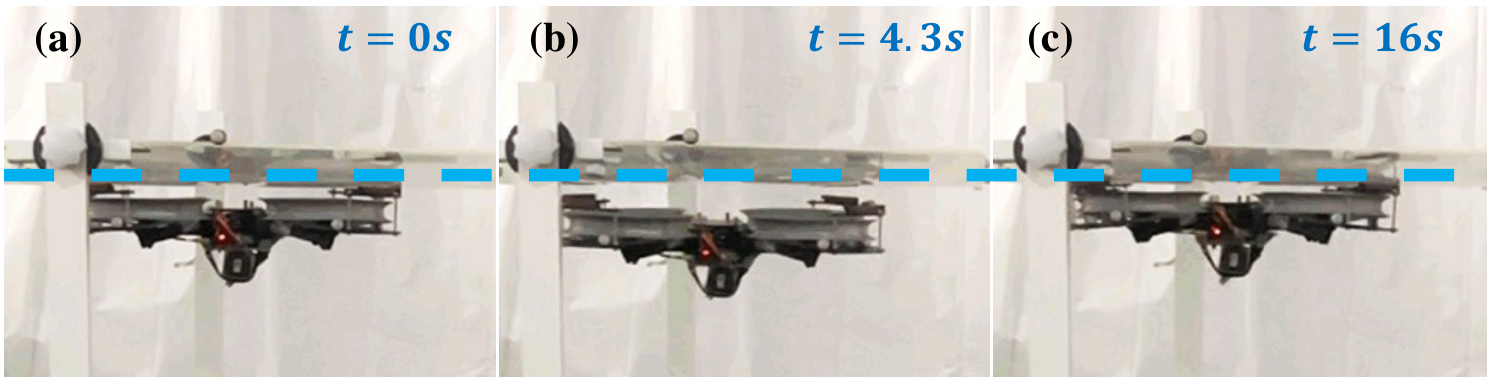}}
 \end{center}
 \caption{The process of the quadrotor attempting power saving.}
 \label{fig:recover}
\end{figure}

\begin{figure}[t]
 \begin{center}
 {\includegraphics[width=1\columnwidth]{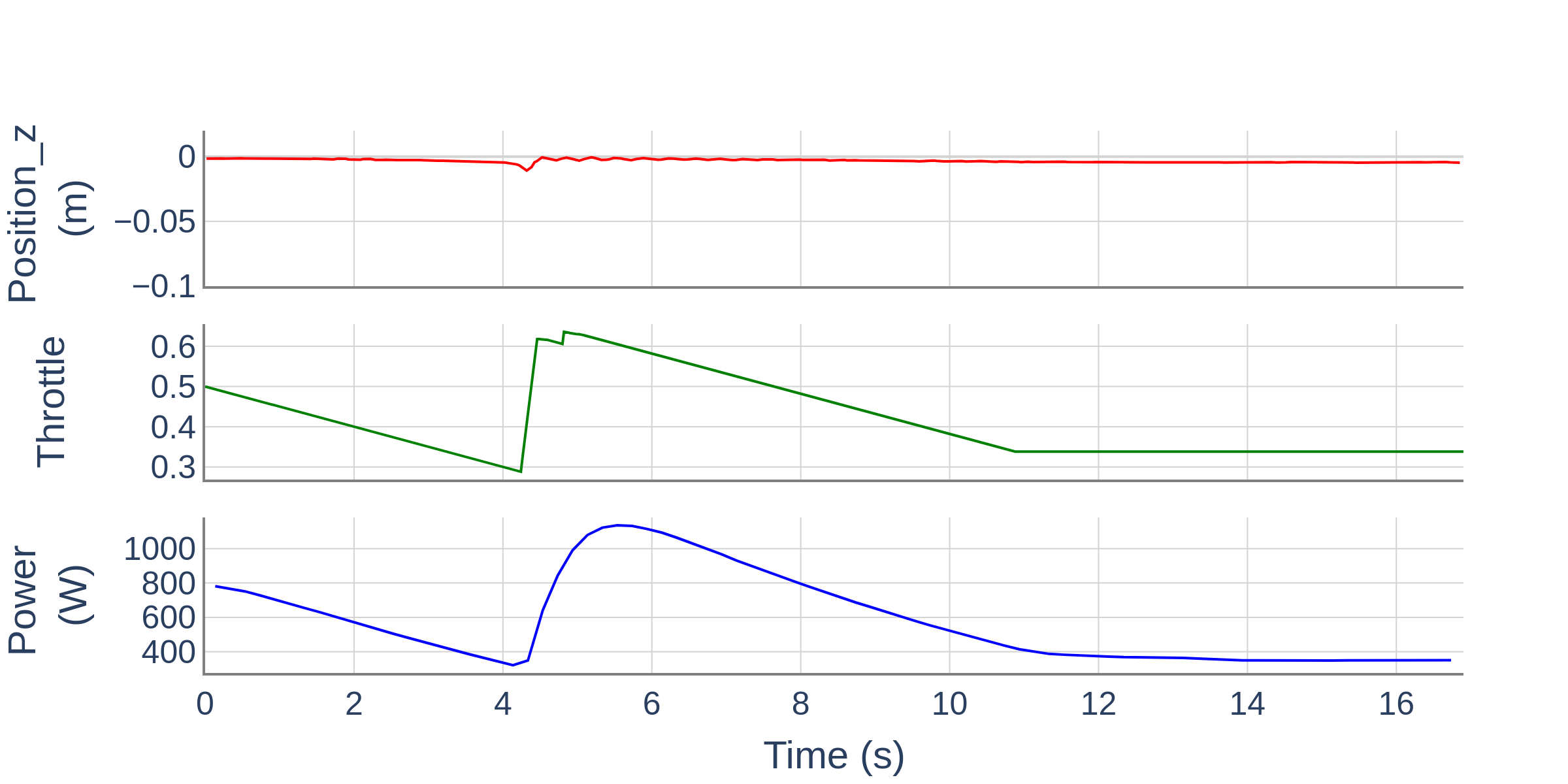}}
 \end{center}
 \caption{The position feedback, throttle command, and power consumption when the quadrotor is attempting power saving.}
 \label{fig:throttle}
\end{figure}

In all perching experiments, we controlled the throttle of the quadrotor based on the methodology mentioned above once it reached the target planes. Fig. \ref{fig:recover} and \ref{fig:throttle} show the process in one of the tests demonstrating perching to the ceiling. As depicted in Fig. \ref{fig:recover} (a), the quadrotor is contacting with the acrylic board. The throttle of to quadrotor is gradually decreased from 0.5 to 0.29 as reported in Fig. \ref{fig:throttle}. The quadrotor drops its altitude at $t=4.3s$ but quickly recovers by increasing the throttle to over 0.6. Then, the throttle is gradually reduced again to 0.34 and maintains the quadrotor's position with an average power of around 340 W. Our method ensures comparatively low power consumption and stable perching of the quadrotor at the same time.

\subsection{Power saving}

\begin{figure}[t]
 \begin{center}
 {\includegraphics[width=1\columnwidth]{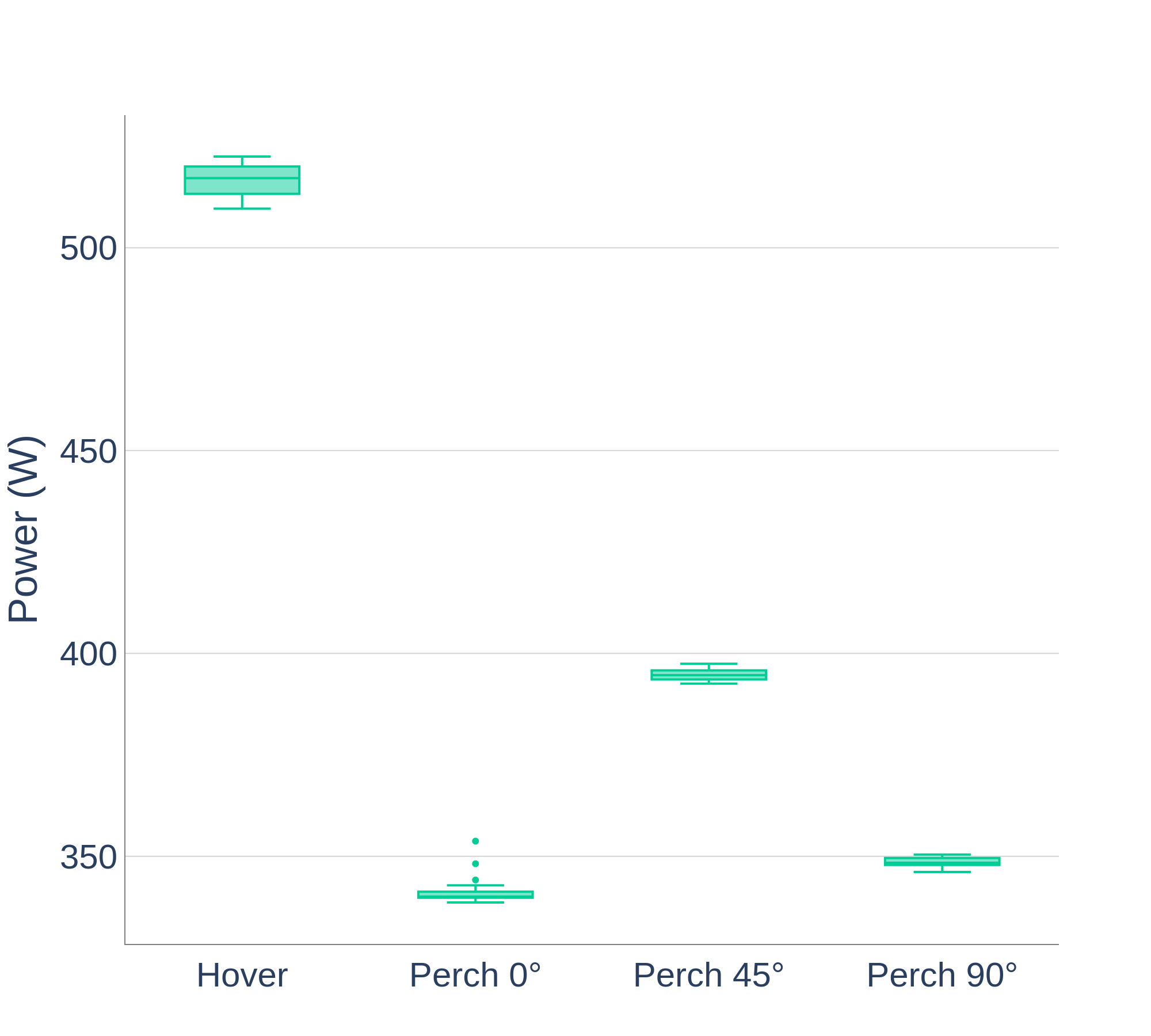}}
 \end{center}
 \caption{Power consumption of the quadrotor in each state.}
 \label{fig:power}
\end{figure}

The power consumption comparison among different states is reported in Fig. \ref{fig:power}. The average power for the quadrotor to hover is 517 W which is gathered when the aircraft is hovering away from the ground, wall, or ceiling. The power consumption data of perching on the planes with incline angles ranging from 0\degree{} to 90\degree{} is obtained when the quadrotor is using throttle equaling to $T_\text{perch}$. The mean power values for perching on 0\degree{}, 45\degree{}, and 90\degree{} planes are 340 W, 394 W, and 348 W, respectively.
Compared to the hovering state, the perching strategy results in energy savings of around 35\%, 24\%, and 33\% in the three perching states accordingly. These results demonstrate the effectiveness of perching by the ceiling effect in improving efficiency while remaining simple and not requiring complicated mechanisms.

Besides, in the cases where the quadrotor perches on planes with 135\degree{} and 180\degree{} incline angles, the rotors can be fully turned off, leading to power consumption close to zero. The novel landing manner is not only easy to achieve but also highly valuable.

\subsection{Stability}
\begin{figure}[t]
 \begin{center}
 {\includegraphics[width=1\columnwidth]{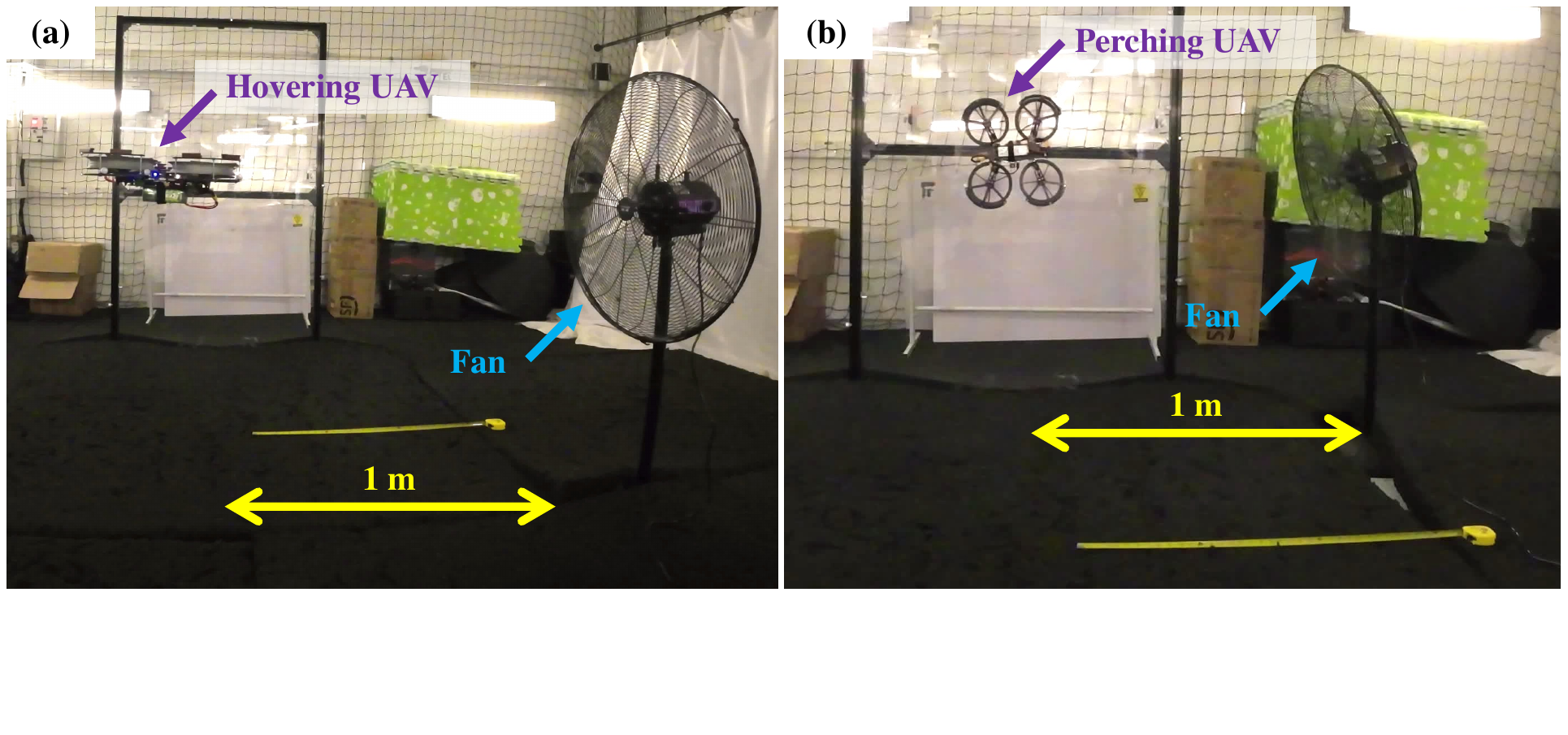}}
 \end{center}
 \caption{The setup for the disturbance tests with gust. (a)Test during hovering. (b) Test during perching}
 \label{fig:wind}
\end{figure}

\begin{figure}[t]
 \begin{center}
 {\includegraphics[width=1\columnwidth]{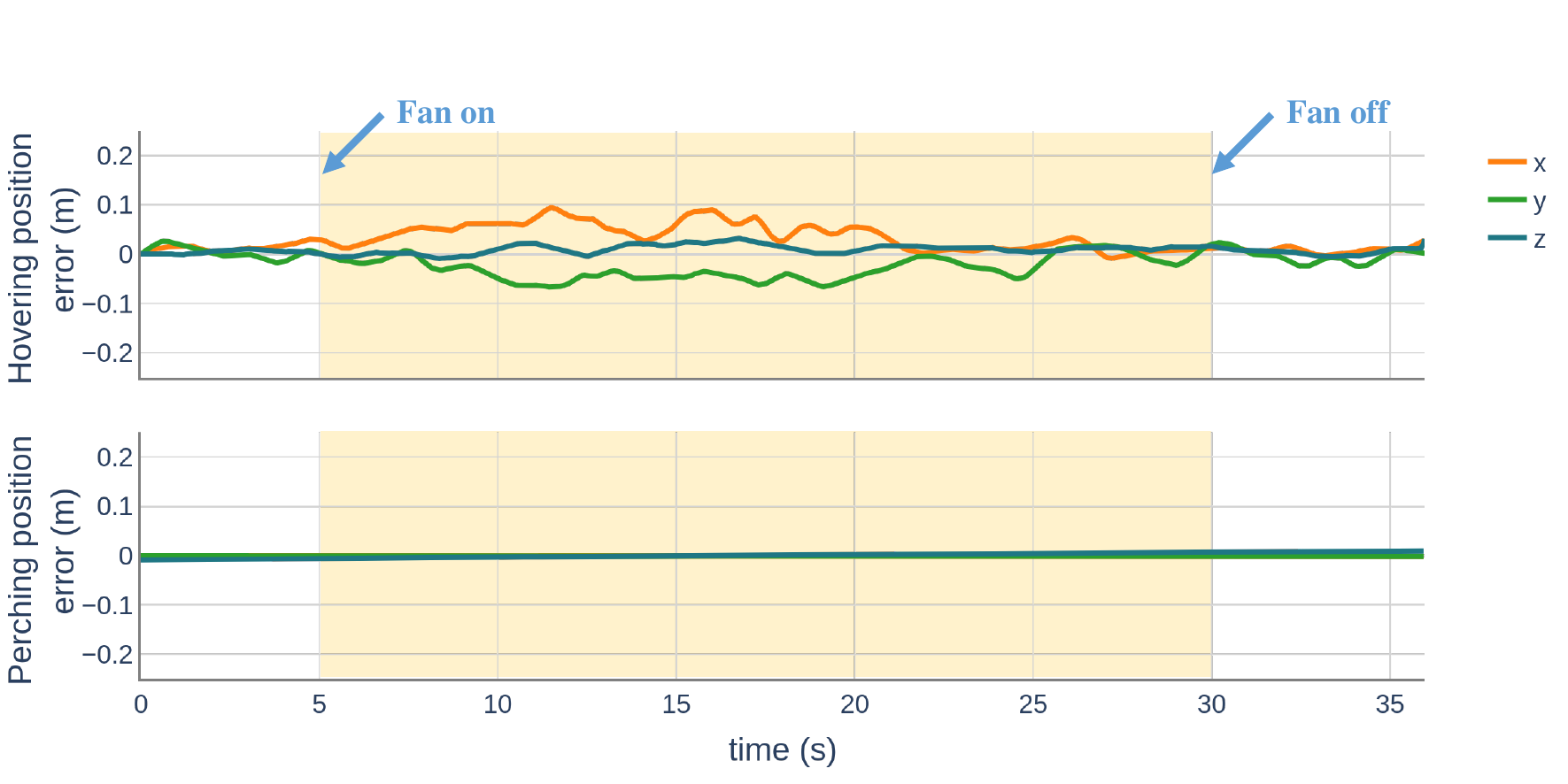}}
 \end{center}
 \caption{The position errors of the quadrotor in the disturbance tests.}
 \label{fig:fan}
\end{figure}

In the disturbance tests, a fan capable of producing a wind speed of 5 m/s is positioned 1 m away from the quadrotor. The detailed setup is displayed in Fig. \ref{fig:wind}. During both the hovering test and the perching test, the fan is turned on for a duration of 25 seconds. Fig. \ref{fig:fan} illustrates that the position errors during hovering are considerably greater in comparison to those during perching, which are all nearly zero. The root-mean-square error is 0.06 m in hovering while it is 0.003 m in perching. Perching demonstrates strong stability and anti-interference ability in this experiment.

\section{Conclusion}
In this work, we proposed to perch a quadrotor on planes by the ceiling effect as a means of saving power and enhancing stability. We designed a quadrotor that can use its propeller guards to make contact with planar structures, thereby eliminating the need for not only landing gear but also grippers, hooks, or adhesive pads. Compared to the existing perching mechanisms of UAVs, our method reduces the complexity of design and is not limited by the angle or material of the perching planes. We have also developed practical perching procedures including trajectory tracking, MPC perching adjustment, and throttle control on surfaces to handle different cases. The power that can be preserved in perching by the ceiling effect is evaluated. Around 30\% of energy consumption can be reduced while guaranteeing the excellent stability of the quadrotor.

Currently, we acknowledge that our quadrotor design may not be optimal in utilizing the ceiling effect. For example, the distance between the propellers and the perching planes is possible to adjust to achieve higher efficiency. Therefore, we plan to further optimize our design and attempt more various scenarios. We will also extend our work to plane detection and autonomous navigation.


\addtolength{\textheight}{-8cm} 



\section*{Acknowledgment}
This work is supported by the Hong Kong Research Grants Council (RGC) General Research Fund (GRF) (no.17206920), the Hong Kong Research Grants Council (RGC) Early Career Scheme (ECS) (no.27202219), and a DJI research donation.

\bibliography{citation} 
\bibliographystyle{IEEEtran}

\end{document}